\documentclass[10pt,conference]{IEEEtran}

\ifCLASSOPTIONcompsoc
  \usepackage[nocompress]{cite}
\else
  \usepackage{cite}
\fi

\usepackage[utf8]{inputenc}
\usepackage{graphicx}
\usepackage{xcolor}

\usepackage{amsmath}
\usepackage{amsfonts}
\usepackage{amssymb}
\usepackage{listings}

\usepackage{balance}
\usepackage{placeins}
\usepackage{hyperref}
\usepackage{multirow}
\usepackage{floatrow}
\begin{document}

\title{Models of Computational Profiles to Study the Likelihood of DNN Metamorphic Test Cases}

\author{
  \IEEEauthorblockN{Ettore Merlo,
    Mira Marhaba,
    Foutse Khomh,
    Houssem Ben Braiek,
    Giuliano Antoniol\\
  }
  \IEEEauthorblockA{Department of Computer and Software Engineering,\\
    Polytechnique Montreal, Montreal, Canada\\
Email: \{ettore.merlo, mira.marhaba, foutse.khomh, houssem.ben-braiek, giuliano.antoniol\}@polymtl.ca}
}

\twocolumn

\maketitle

\thispagestyle{plain}
\pagestyle{plain}

\begin{abstract}
Neural network test cases are meant to exercise different reasoning paths in an architecture and used to validate the prediction outcomes.

In this paper, we introduce ``computational profiles'' as vectors of neuron activation levels. We investigate the distribution of computational profile likelihood of metamorphic test cases with respect to the likelihood distributions of training, test and error control cases. We estimate the non-parametric probability densities of neuron activation levels for each distinct output class. Probabilities are inferred using training cases only, without any additional knowledge about metamorphic test cases.
 
Experiments are performed by training a network on the MNIST Fashion library of images and comparing prediction likelihoods with those obtained from error control-data and from metamorphic test cases.

Experimental results show that the distributions of computational profile likelihood for training and test cases are somehow similar, while the distribution of the random-noise control-data is always remarkably lower than the observed one for the training and testing sets.

In contrast, metamorphic test cases show a prediction likelihood that lies in an extended range with respect to training, tests, and random noise. Moreover, the presented approach allows the independent assessment of different training classes and experiments to show that some of the classes are more sensitive to misclassifying metamorphic test cases than other classes.

In conclusion, metamorphic test cases represent very aggressive tests for neural network architectures. Furthermore, since metamorphic test cases force a network to misclassify those inputs whose likelihood is similar to that of training cases, they could also be considered as adversarial attacks that evade defenses based on computational profile likelihood evaluation.

\end{abstract}

\IEEEpeerreviewmaketitle

\section{Introduction}

Deep neural networks (DNNs) are more and more integrated in large and industrial software systems in many fields. Metamorphic testing \cite{metamorph_review_2018, deepevolution, xie2019deephunter} can be used to create neural network test cases based on generated variations from the original data using a variety of mutations and transformations.

We describe a statistical approach to measure the ``reasoning'' likelihood during a network prediction. Motivation for reasoning likelihood measurement comes from the intuition that a network's precision and performance observed during training and tests cannot be assumed to be the same when dealing with cases that traverse very different computational paths in the network during prediction. Indeed, we don't know the a-priori precision and performance of a network in these cases. Sometimes the prediction is correct, but very often it's incorrect, especially for ``corner cases'' that are tests specifically designed to test rare and unusual events and that may cause a network to fail.

DNN-based software systems are vulnerable to ``unusual'' and ``unexpected'' corner cases that may cause an incorrect network prediction. Sensitive and critical domains, such as aerospace, medicine, finance, or cyber-security require a higher confidence in the DNN prediction and more robustness against corner cases.

Adversarial attacks can be considered as particular corner cases that have been specifically designed to make a network fail. Recent research has been devoted to the detection of adversarial cases, when it comes to neural networks for image recognition \cite{metzen_iclr_2017, tramer_iclr_2018, tramer_icml_2020, yuan_tse_2019}. Good defense results have been obtained by somehow learning adversarial features and behaviors and using this knowledge to distinguish between correctly and incorrectly predicted classifications due to adversarial attacks.

Approaches involving extracting computational information from a network during classification and using it for adversarial cases detection have been reported in the literature \cite{papernot_distillation_arxiv_2015, surprise_icse_2019, raid_arxiv_2020}. These approaches rely on extracting computational information from a network and then on somehow training a secondary classifier using adversarial features and behavior to distinguish between correctly and incorrectly predicted classifications due to adversarial attacks.

We define an approach based on the estimation of non-parametric probability densities of neuron activation levels during training of each output class only, without using secondary training of an additional classifier and without any knowledge of corner cases. We define the extracted vectors of neuron activation levels at the different layers as the ``computational profile'' of an input during processing. Computational profiles can be extracted during training and also during classification. During classification, we compute the likelihood that an observed computational profile based on neuron activation levels is produced by the model corresponding to the best predicted class. The approach is presented in Section \ref{lab-method}.

DNN training has been performed on images from the MNIST-fashion database \cite{MNIST-fashion} and neuron activation level probabilities have been estimated during training.

Experiments have been performed measuring the prediction likelihood on images composed of random pixels as error control-data and on metamorphically transformed images by rotation. Results are presented in Section \ref{sec-results} and show that the distributions of computational profile likelihood for training and test cases are somehow similar, while the distribution of the random-noise control-data is always remarkably lower than that observed for the training and testing sets.

In contrast, metamorphic test cases show a prediction likelihood that lies in an extended range  with respect to training, tests, and noise. Also, the presented approach allows the independent assessment of different training classes and experiments show that some of the classes are more sensitive to misclassifying metamorphic test cases than other classes.

Research questions can be expressed as follows:
\begin{itemize}
\item RQ1: What are the computational profile likelihood distributions of training, test, error control-data and metamorphic test cases?
\item RQ2: How are these distributions similar or different from a statistical analysis' point of view?
\end{itemize}

Contributions of this paper can be summarized as follows:
\begin{itemize}
\item the experimental measure of likelihood distribution for training, test, error control-data and metamorphic test cases on MINST-fashion database of images
\item the statistical analysis comparison of these distributions
\end{itemize}

Future research perspectives are presented in Section \ref{lab-future-res}, while related works and threats to validity are discussed in Section \ref{lab-rel-work} and Section \ref{lab-ttv}, respectively.

\section{Data Derivation}
\label{lab-data-deriv}

In this section, we define the distorted data that have been deliberately derived from the original data using a variety of mutations and transformations. Indeed, the original data belonging to the MNIST-fashion database of images have been divided into training data, validation data and testing data to serve the multiple phases of model development and evaluation.

\subsection{Pixel-value Mutations}
Intuitively, the pixel-value mutations aim to change the pixel values of the original image.

In principle, we could have perturbated the pixels using small, even imperceptible random distortions that constrain the deviation induced between the mutated and original images within a user-configurable $L_2$-norm. Regarding the image class, this type of constrained mutation allows to attribute the same class as original input to the mutated one.

For our experiments, rather than slightly perturbing the images, we completely mutated the pixels using random noise with no constraints on the difference between the mutated element and the original one. This pure random mutation will generate images of the same size as the original data (i.e. size refers to height, width, and depth) and containing random pixel values that lie in the same range as the original data used to train the subject neural network. All the corresponding model outputs would be considered as wrong predictions, and so these images are considered as error control-data for comparison with the other datasets.

\subsection{Affine Transformations}
\label{lab-affine-tranf}

While pixel-value mutations change the pixels’ intensities, affine transformations move the pixels of the image while somehow maintaining the pixel intensity. First of all, this allows to have more variability within the data inputs included in the study and the choice of affine transformation was motivated by the research\cite{logan2018rotation} showing that performing rotations and translations alone can be used to completely fool an image classifier, even when the latter is robust against the $L_2$-norm-bounded adversary. In our work, we consider mainly three types of affine transformations as described in the Table \ref{tab:affine_trans} below. 

\begin{table}[htbp]
\centering
\caption{Affine Data Transformations and Parameters}
\label{tab:affine_trans}
\begin{tabular}{|l|l|l|}
\hline
Name                  & Params    & Short Description  \\
\hline
Translation     & ($x$,$y$) & it shifts the object by the \\
&           & provided coordinate steps.\\
\hline
Scaling     & ($x$,$y$) & it can shrink or zoom the size \\
&           & of an object along the axis.\\
\hline
Rotation   & $\alpha$   & it rotates the image by $\alpha$ degrees. \\
\hline
\end{tabular}
\end{table}

Then, we create multiple derived datasets based on those transformations. On the first hand, we aim to define the appropriate range of values that preserve the data semantics, so the produced input can inherit the label of its parent input as well. Given the original data, we manually set up a wide range of values, then we narrow down the range as we progress by a trial and error process that consists of the following steps: (1) we select a transformation and a subset of data inputs; (2) we sample, repeatedly, random parameters’ values from the prefixed range; (3) we visually check the transformed inputs to check if they lose their meaning or deviate significantly from the original images; (4) we reduce the range of parameters' values to re-iterate the process. The tuning process is terminated when we correctly define two bounds that represent high and low accepted values for each transformation. Thus, we consider the resulting constrained transformations as semantically-preserving transformations. Formally, given an image $I$, the application of a semantically-preserving transformation on $I$ generates another new image $I'$ such that the semantics of $I$ and $I'$ are the same from the human perspective. Like the images resulting from constrained perturbations, we can divide the produced images from semantically-preserving transformations into two datasets: benign data for which the model still recognizes its label and adversarial data that is misclassified. In the case of this paper, we study the profiles of only the latter of the two datasets. In other words, we keep only the adversarial (misclassified) data.

\subsection{Characteristics of the Derived data}
\begin{figure*}[ht]
\centering
\includegraphics[scale=0.75]{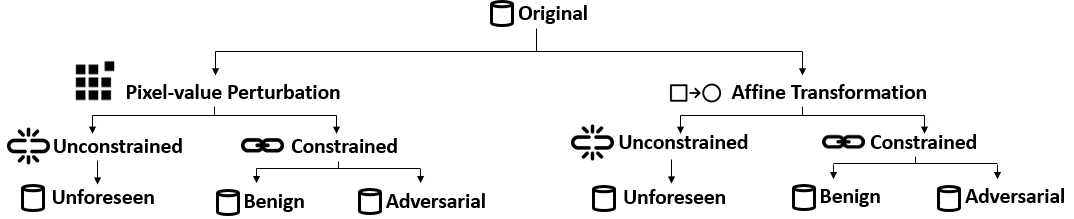}
\caption{Overview of Derived Data Map}
\label{fig:overview_data_map}
\end{figure*}

Adversarial examples are well studied in the literature\cite{biggio2018wild} because of their high potential in detecting vulnerabilities and highly-sensitive models. The semantically-preserving transformations are also exploited in \cite{deepevolution, xie2019deephunter} with the objective of applying metamorphic testing, where the model should predict the same label for both original and transformed inputs to pass the test. As illustrated in our derived data map in Figure \ref{fig:overview_data_map}, we consider all the previous transformed data and we even push further towards unforeseen data that can be only resulting from malicious hand-crafted attacks that challenge the security of any DNN-based software systems exposed to users via websites or mobile phone apps.

Presented experiments have been performed using random pixel-value mutations as error control-data and rotations as constrained affine transformations. Affine parameters have been selected to make a network misclassify all rotated inputs into a best predicted class that was different from the original one prior to transformations.

\section{Method}
\label{lab-method}

A DNN is typically composed of several layers of neurons that carry and propagate activation levels across layers using weights and transfer functions\cite{liu_survey_dnns_2017}.

Previous approaches extracted behavior information from the network during classification and use it to train a secondary classifier \cite{papernot_distillation_arxiv_2015, surprise_icse_2019, raid_arxiv_2020}.

\subsection{A non-parametric approach}
\label{lab-non-param}

Some previous stochastic approaches suffered from the assumption of normal distribution hypothesis  \cite{papernot_distillation_arxiv_2015, raid_arxiv_2020}. Indeed, observed distributions of activation levels are not normal based on the results of statistical tests we have performed \cite{scipi_stats}. Furthermore, when the RELU transfer function was used, we observed distributions that are highly skewed and highly sparse, in contrast with the normality hypothesis. The use of Gaussian distribution is not appropriate for modeling the density of activation levels because it would result in poor fitting.

Since the shape of the probability density of the activation levels is unknown and we didn't want to make explicit assumptions, we used a non-parametric probability estimation approach to overcome previous limitations.

In this context, we estimated the probability density functions $P(i, j, x, k)$ of the activation level of a neuron $i$ in layer $j$, caused by the processing of an input vector $x$ from the training set under the hypothesis of $x$ belonging to class $k$.

Differently from a similar research work \cite{neural_prob_zheng_nips_2018} that used Gaussian mixtures to estimate the probability densities of neuron activation levels, we used histograms for the same purpose and as an alternative method, because of the very highly skewed distribution of probability densities observed in our data.

$nodeAvgLev(i, j, X, k)$ and $nodeStdVar(i, j, X, k)$ computed in Equations \ref{eq-node-avg} and \ref{eq-node-st-dev} are respectively the average and the standard deviation of the distribution of activation levels of the $i$-th neuron in the $j$-th layer, for inputs $x$ in the training set $X$, belonging to class $k$ of the set of all classes $K$.

\begin{equation}
  \begin{array}{c}
    nodeAvgLev(i, j, X, k) = \displaystyle{\frac{1}{\mid X \mid}} \cdot \displaystyle{\sum_{x \in X}} actLev(i, j, x, k)\\
  \end{array}
  \label{eq-node-avg}
\end{equation}

\begin{equation}
  \begin{array}{l}
    nodeStdVar(i, j, X, k) = \\
    \\
  \displaystyle{\frac{1}{\mid X \mid}} \cdot \sqrt{ \displaystyle{\sum_{x \in X}} (actLev(i, j, x, k) - nodeAvgLev(i, l, X, k))^2}
  \end{array}
  \label{eq-node-st-dev}
\end{equation}

The resolution of distribution estimation is determined by the width of histogram slots. To have a representative resolution across the neuron distributions, we used bins of variable size $width(i, j, X, k)$ for a neuron $(i, j)$, as follows:

\begin{equation}
  \begin{array}{c}
    width(i, j, X, k) = c * nodeStdVar(i, j, X, k)
  \end{array}
  \label{eq-hist-width}
\end{equation}

In the presented experiments, $c ~= ~1$ has been used, but $c ~= ~0.5$ could be used for a finer estimation of distributions.

The bin identifier $binId(i, j, x, k)$ corresponding to an activation level $actLev(i, j, x, k)$ of a neuron $(i, j)$ and an input $x$ belonging to class $k$ in the training set $X$ is:

\begin{equation}
  \begin{array}{c}
    binId(i, j, x, k) = int \left ( \displaystyle
    \frac{actLev(i, j, x, k)}{width(i, j, X, k)} \right )
  \end{array}
\end{equation}

Bin frequencies $bFreq(b, i, j, X, k)$ have been computed during training by counting how many times inputs from the training set $X$ and belonging to class $k$ have produced an activation level that falls into bin identifier $b$ for neuron $(i, j)$.

Bin probabilities have been computed from bin frequencies as follows:

\begin{equation}
  \begin{array}{c}
    p(b, i, j, X, k) = \displaystyle{\frac{1}{\mid X \mid \cdot \mid K \mid}} \cdot bFreq(b, i, j, X, k)\\
  \end{array}
  \label{eq-bin-prob}
\end{equation}

To smooth probabilities over the bins, a very low probability has nevertheless been assigned to all bins with null frequencies.

The estimated likelihood $L(y, j, k, X)$ of the activation levels of neurons in layer $j$, for the input $y$ under the hypothesis of $x$ belonging to class $k$, has been computed as the joint probability of all neurons composing layer $j$. The joint probability is obtained as indicated in Equation \ref{eq-prob:a} by multiplying together the bin probabilities $p(b, i, j, X, k)$ from Equation \ref{eq-bin-prob}. Since the joint probabilities are quite small, it is practical to convert them into logarithmic values, as shown in Equation \ref{eq-prob:b}. By using the properties of logarithms, multiplications can be replaced by additions as reported in Equation \ref{eq-prob:c}. Logarithmic probabilities are negative or equal to zero real numbers. We define $dist(y, j, k, X)$ as the symmetric positive value of logarithmic probabilities, as reported in Equation \ref{eq-prob:d}. The final formula for the distance $dist(y, j, k, X)$ used for the presented experiments is then reported in Equation \ref{eq-prob:e}.

High distance values are related to low probabilities, while small distances are related to high probabilties. So, a small distance value means that the input is likely to present a computational profile close to those observed during training for the same class. On the other hand, a high distance for an input says that the probability of its computational profile with respect to the those observed during training is low.

\begin{subequations}

  \begin{equation}
  \begin{array}{c}
    L(y, j, k, X) = ~\displaystyle{\prod_{i}} ~p(b, i, j, X, k)\\
    \\
  \end{array}
  \label{eq-prob:a}
\end{equation}

\begin{equation}
  \begin{array}{c}
    log(L(y, j, k, X)) = log \left ( ~\displaystyle{\prod_{i}} ~p(b, i, j, X, k) \right )\\
    \\
  \end{array}
  \label{eq-prob:b}
\end{equation}

\begin{equation}
  \begin{array}{c}
    log(L(y, j, k, X)) = ~\displaystyle{\sum_{i}} log( ~p(b, i, j, X, k))\\
    \\
  \end{array}
  \label{eq-prob:c}
\end{equation}

\begin{equation}
  \begin{array}{c}
       dist(y, j, k, X) = - log(L(y, j, k, X))\\
 \\
      \end{array}
  \label{eq-prob:d}
\end{equation}

\begin{equation}
  \begin{array}{c}
    dist(y, j, k, X) = - ~\displaystyle{\sum_{i}} ~log(p(b, i, j, X, k))\\
  \end{array}
  \label{eq-prob:e}
\end{equation}

\end{subequations}

In the presented experiments, we concentrated on the last but one layer only and results have been obtained using distances $dist(y, N-1, k, X)$, where $N$ is the number of layers in a network and $(N - 1)$ is the layer before the output layer. A brief discussion about this choice is presented in section \ref{lab-ttv}.

\subsection{Statistical analysis}
\label{lab-stats}

Summary statistics, including averages and ranges, were calculated for each class. They are reported in Tables \ref{tab-train-dist-params}, \ref{tab-test-dist-params}, \ref{tab-rnd-dist-params}, and \ref{tab-rot3-dist-params}.

To compare distances and probabilities over classes, data exploration and visualization followed by statistical analysis of random-pixel and misclassified metamorphically mutated images have been applied. Statistical tests were performed using the Statistical Functions library ``scipy.stats'' \cite{scipi_stats}.
Their results are reported in Tables \ref{tab:stat-train-test}, \ref{tab:stat-train-rnd}, and \ref{tab:stat-train-rot3}.

We have reported results about Epps-Singleton two sample test and Cliff’s delta. Results for Cliff's delta have been computed using Ernst's Python implementation \cite{ernst_cliffs_delta} that is derived from Torchiano's in R \cite{marco_torchiano_effect_size}. In addition, we have also performed the Wilcoxon rank sum test for continuous and discrete variables, the Kolmogorov-Smirnov, and the Anderson-Darling tests. Their results were similar and consistent with the Epps-Singleton two sample test and have not been reported, because of being somewhat redundant.

We also studied Cliff’s delta and Cohen-d effect size as non parametric tests \cite{hess_2004, romano_2006}.  Results for Cohen-d have been implemented by ourselves.
Cliff’s delta and Cohen-d effect size are non parametric tests \cite{hess_2004, romano_2006}.
Cohen's d results were similar and consistent with Cliff's delta and have not been reported.

For Cliff's delta $d$, the magnitude is assessed using thresholds on $d$ and is classified as follows: "negligible", if $d < 0.147$;  "small", if $d < 0.33$; "medium", if $d < 0.474$; and "large'' otherwise \cite{romano_2006}.

The  alpha level is  the chance  taken by  researchers to  make a  type one error.  The  type  one  error  is  the error  of  incorrectly  declaring  a difference, effect  or relationship  to be true  due to chance  producing a particular state of events. The alpha level for one test can, in principle, be set at 0.05; in other words,  in  no  more than  one  in  twenty statistical  tests  will the  test  show ``something'' while  in fact there is nothing.  Since we carried out more than one statistical  test, the  chance of  finding at  least one  test statistically significant  due  to  chance  fluctuation,  and to  incorrectly  declare  a difference or relationship to be  true, increases.

We hypothesize that there is a substantial difference between adversarial attacks and legitimate images for the different classes.  In other words we formulated 10 (one per class)  hypotheses and thus statistical tests should be corrected accordingly. Consequently, conservatively applying Bonferroni correction, we assumed an alpha value of 0.005, instead of 0.05. Test results were corrected with the Benjamini \& Hochberg procedure \cite{benjamini_1995}.

Following this first decision step, if needed, we applied the Epps-Singleton test to further verify consistency of findings.

\section{Experiments and Results}
\label{sec-results}

\subsection{Image Sets}
The experiments in this paper take four image sets into account: train, test, random-pixel, and rotated images.

\subsubsection{Train and Test Images}
For the first two image sets, we decided to use the MNIST Fashion library of 70000 images \cite{MNIST-fashion}, and split them into 60000 training images and 10000 testing images.

\subsubsection{Images with Random Pixels}
As some sort of error control-data, 60000 images of the same MNIST size were generated by creating completely random values for each pixel, using a uniform distribution. Each row's pixels were randomly generated in sequence, for each image, without re-initializing the random generator between rows or between images. Although these images are best classified by the network into some classes, we assumed that all these predictions be errors.

\subsubsection{Rotated Images}
This last image set allows to analyze the activation values of images that the network believes that it recognized, but that have in fact been perturbed. Starting from the same 60000 training images, we perform affine transformations repeatedly until the point that the image prediction goes from correct to incorrect. In the case of this article, we study the effect of rotation on the images that were incorrectly classified by the model after rotation.

\begin{figure}[htbp]
\setlength\tabcolsep{0pt}
\renewcommand{\arraystretch}{0}
\begin{center}
\begin{tabular}{lcc}
  \includegraphics[keepaspectratio=1,width=0.5\linewidth]{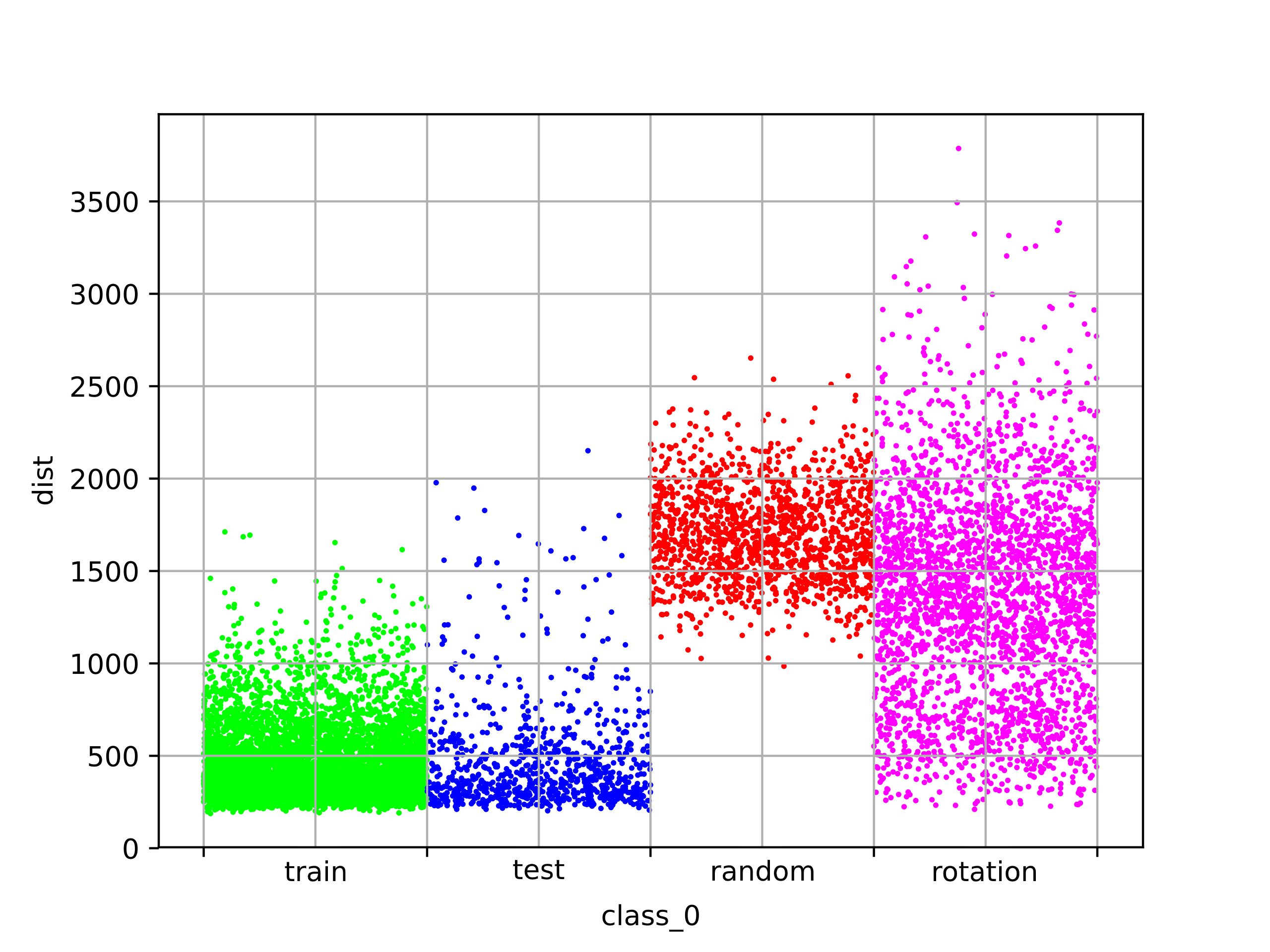} &
  \includegraphics[keepaspectratio=1,width=0.5\linewidth]{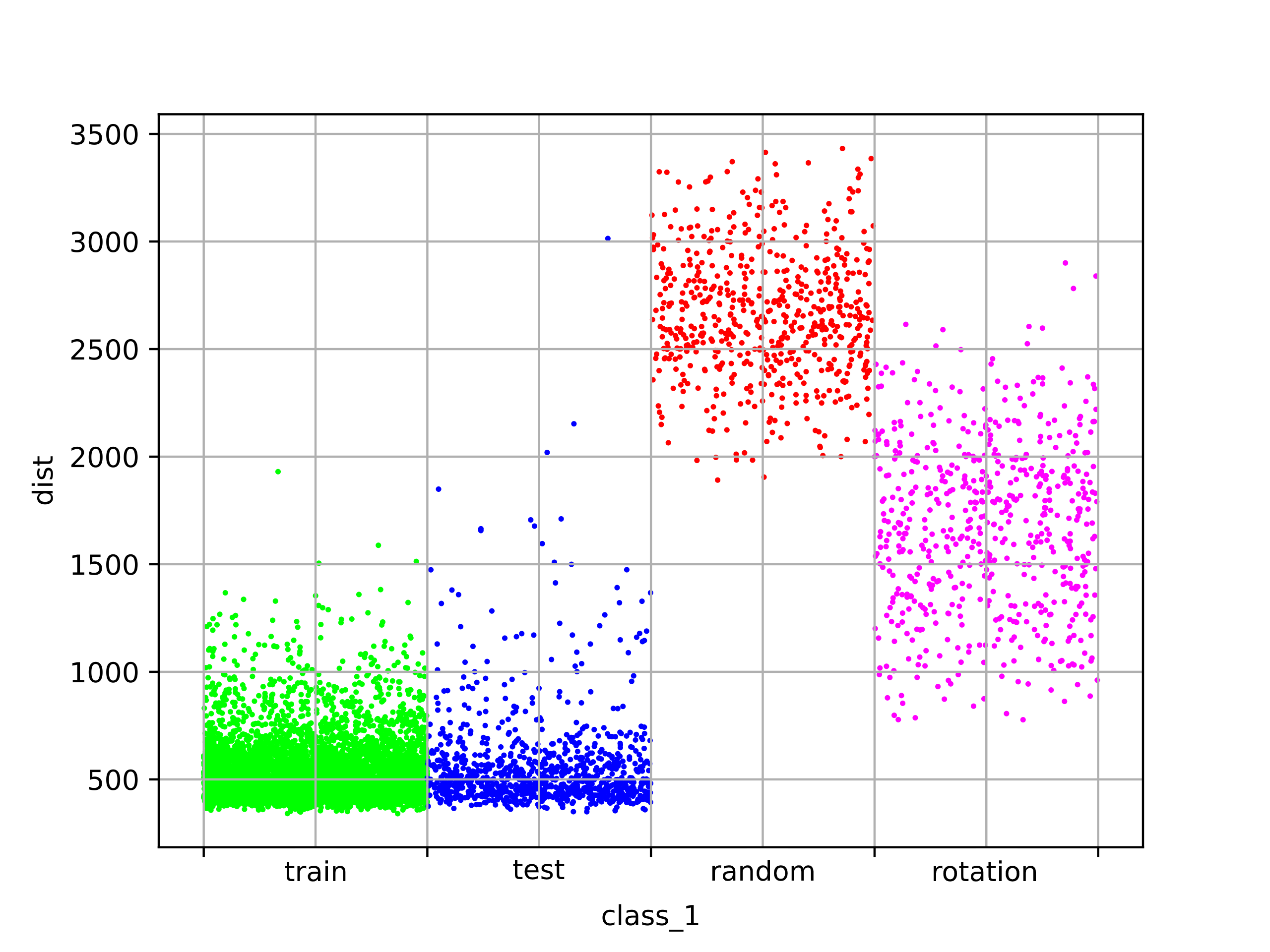}\\
  \includegraphics[keepaspectratio=1,width=0.5\linewidth]{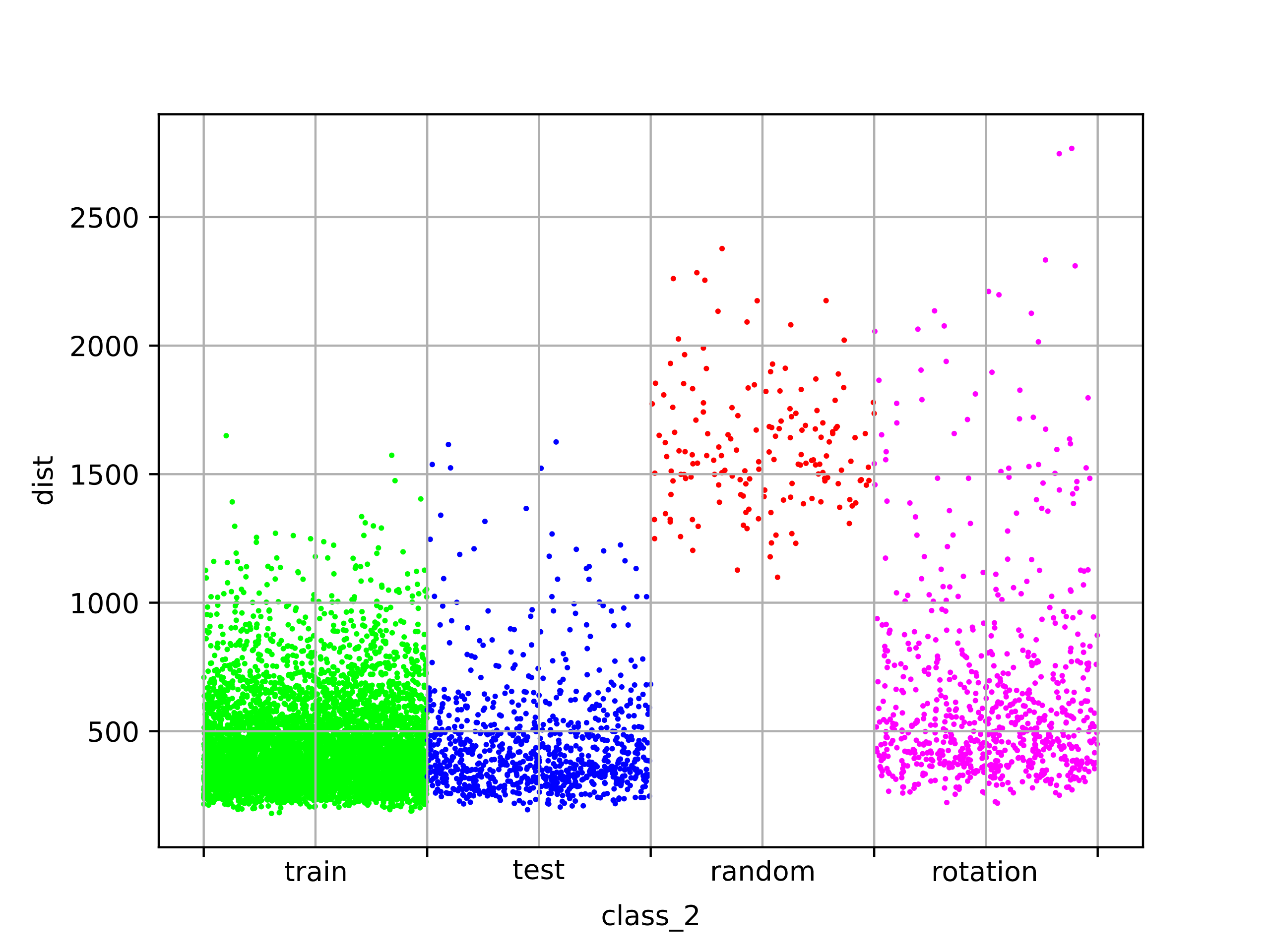} &
  \includegraphics[keepaspectratio=1,width=0.5\linewidth]{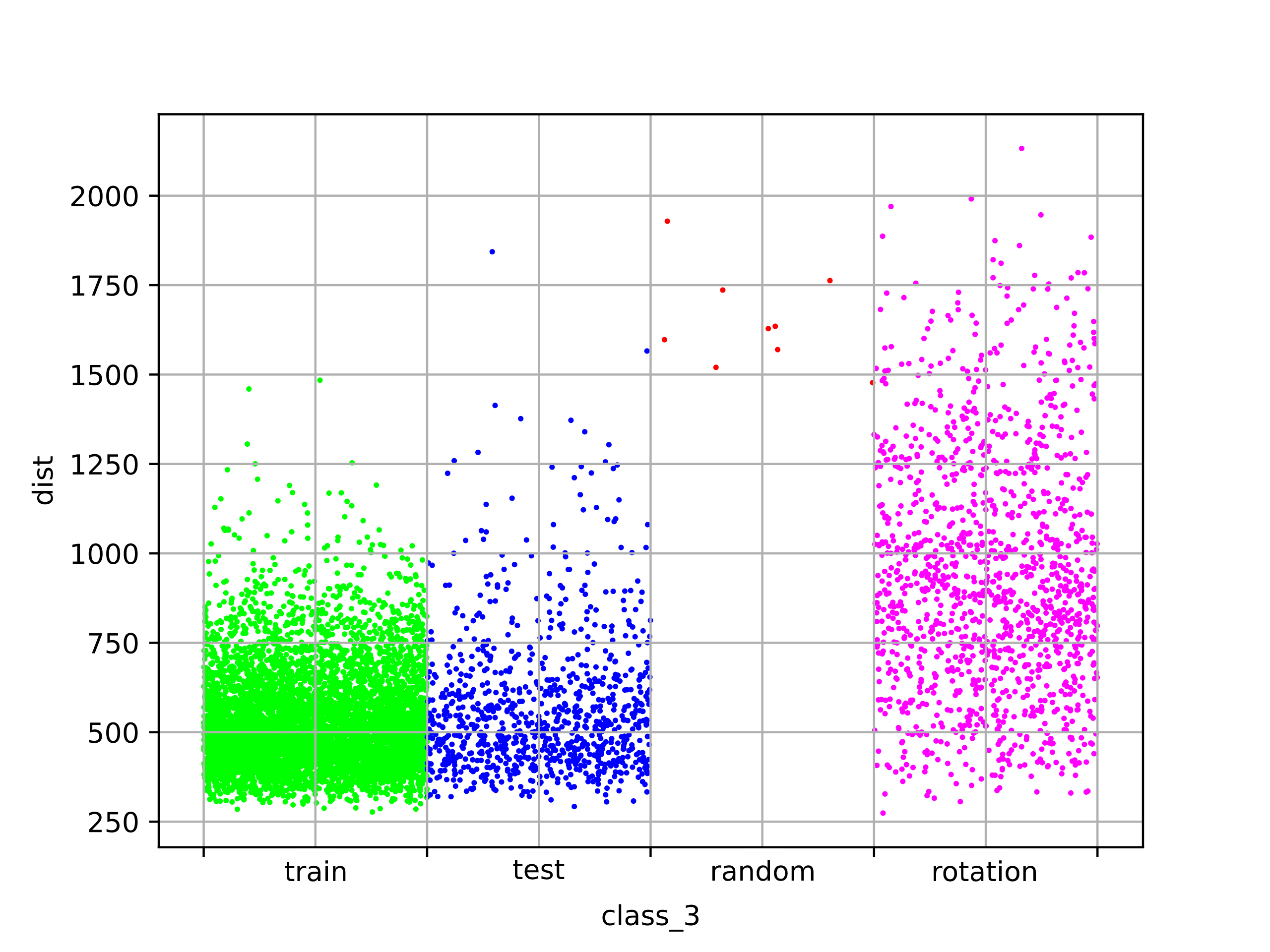}\\
  \includegraphics[keepaspectratio=1,width=0.5\linewidth]{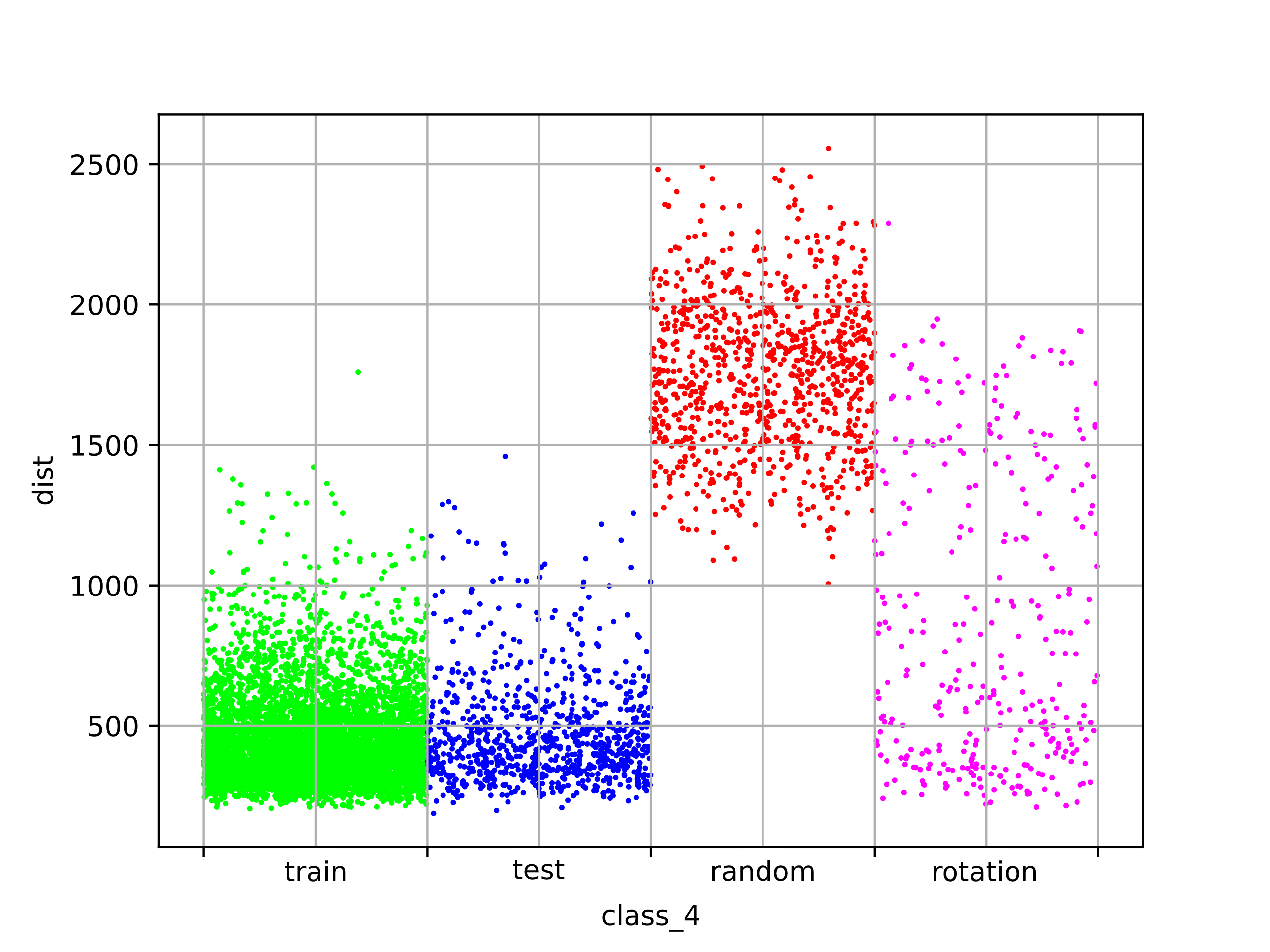} &
  \includegraphics[keepaspectratio=1,width=0.5\linewidth]{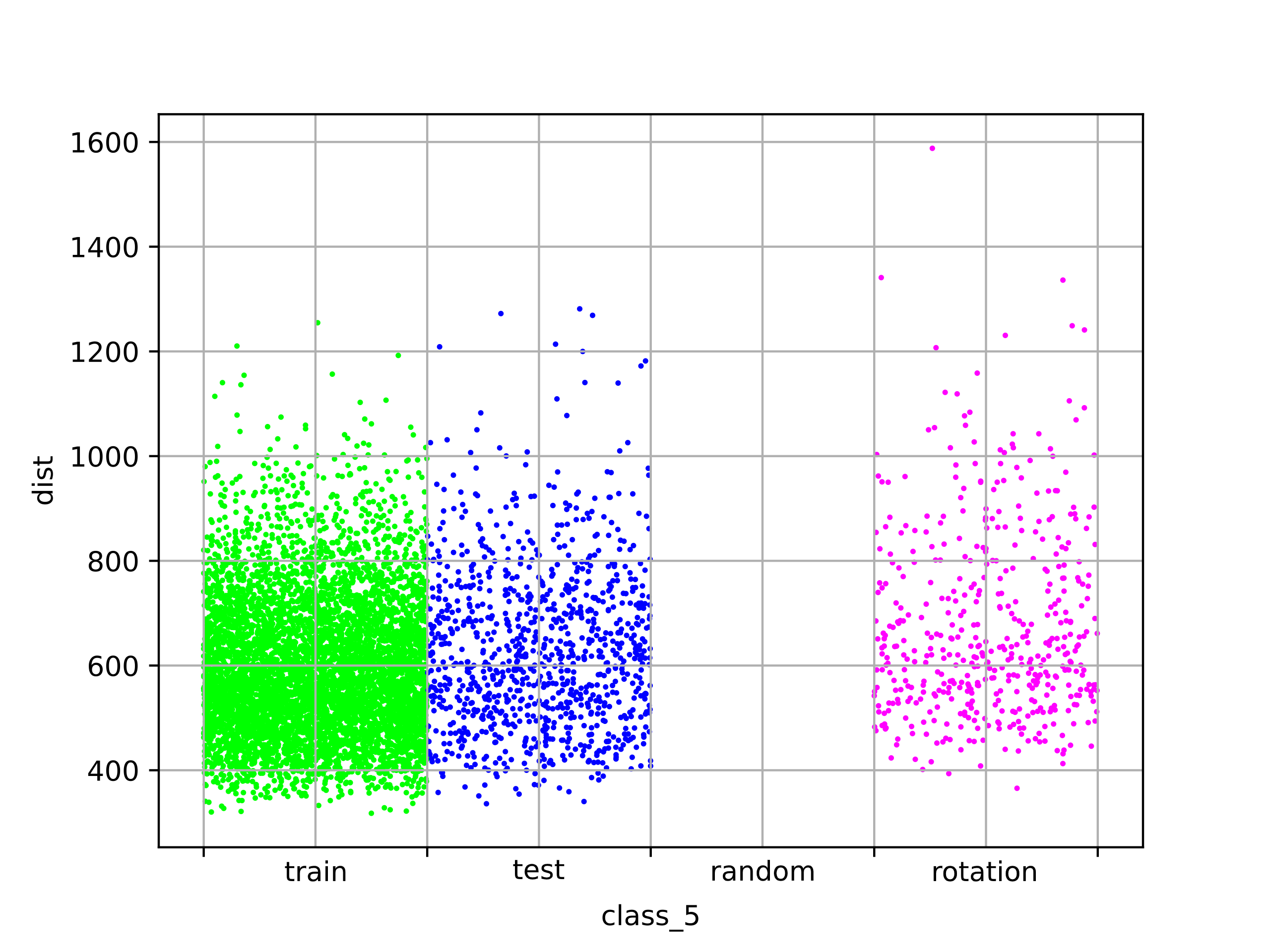}\\
  \includegraphics[keepaspectratio=1,width=0.5\linewidth]{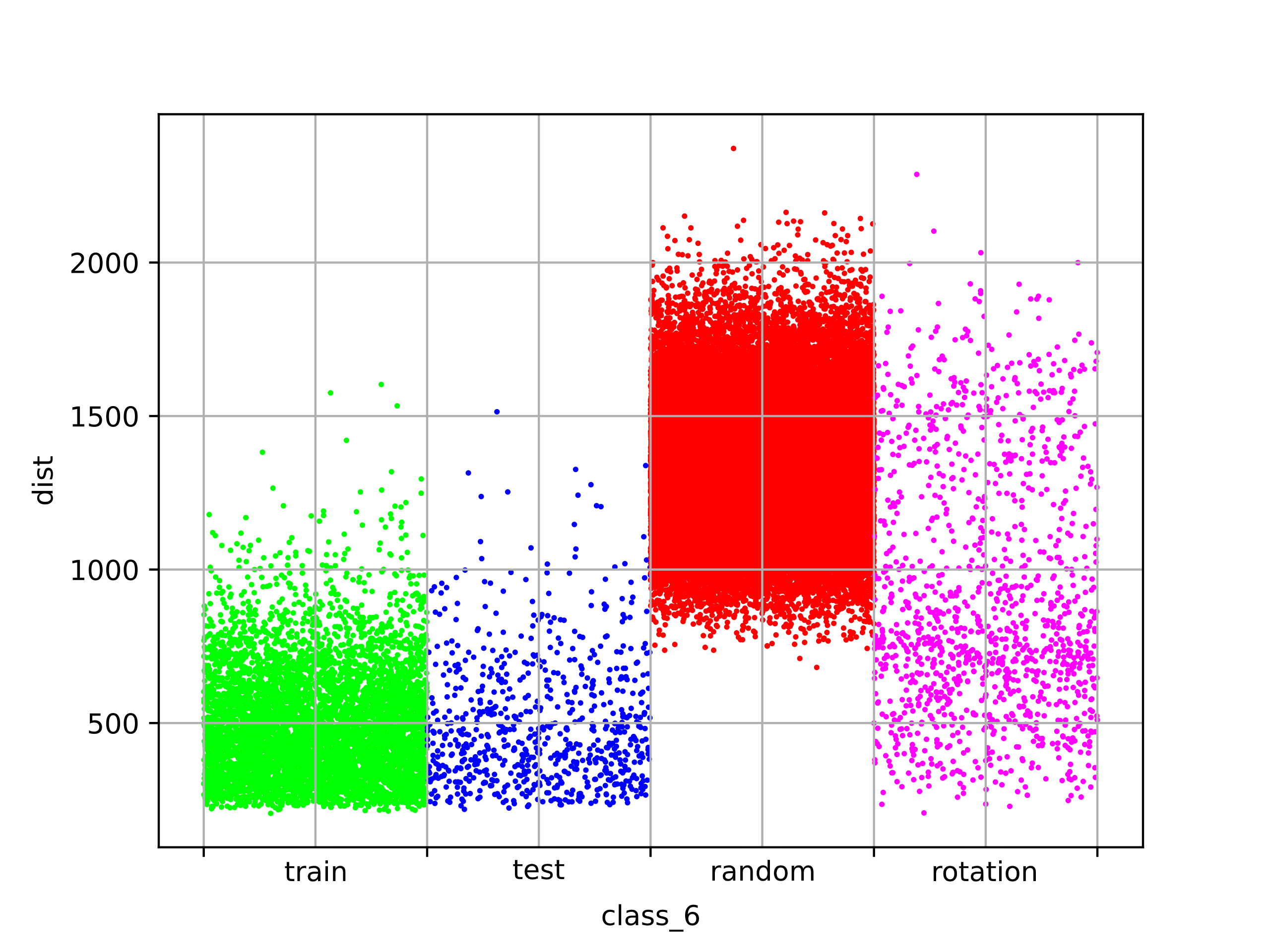} &
  \includegraphics[keepaspectratio=1,width=0.5\linewidth]{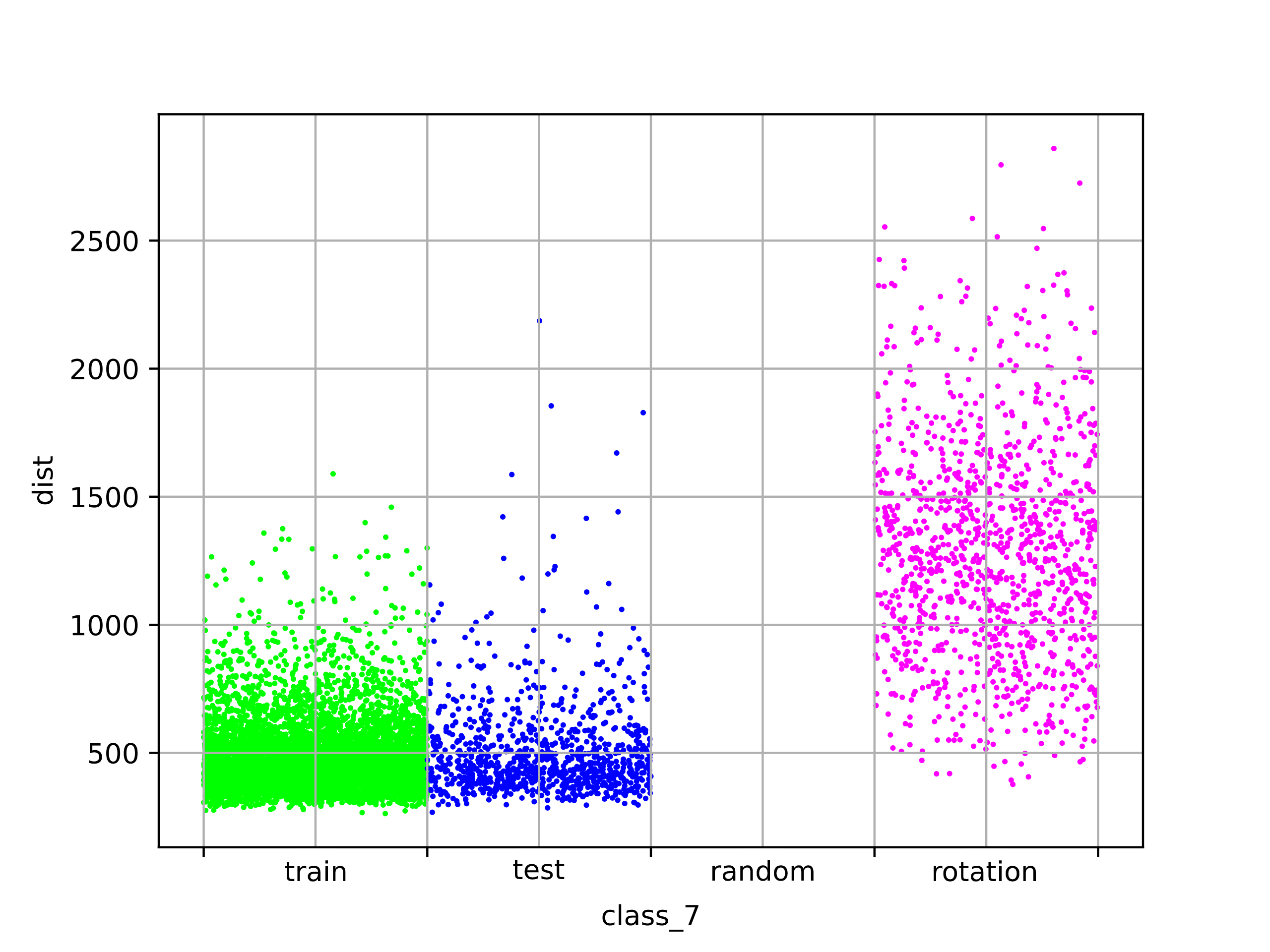}\\
  \includegraphics[keepaspectratio=1,width=0.5\linewidth]{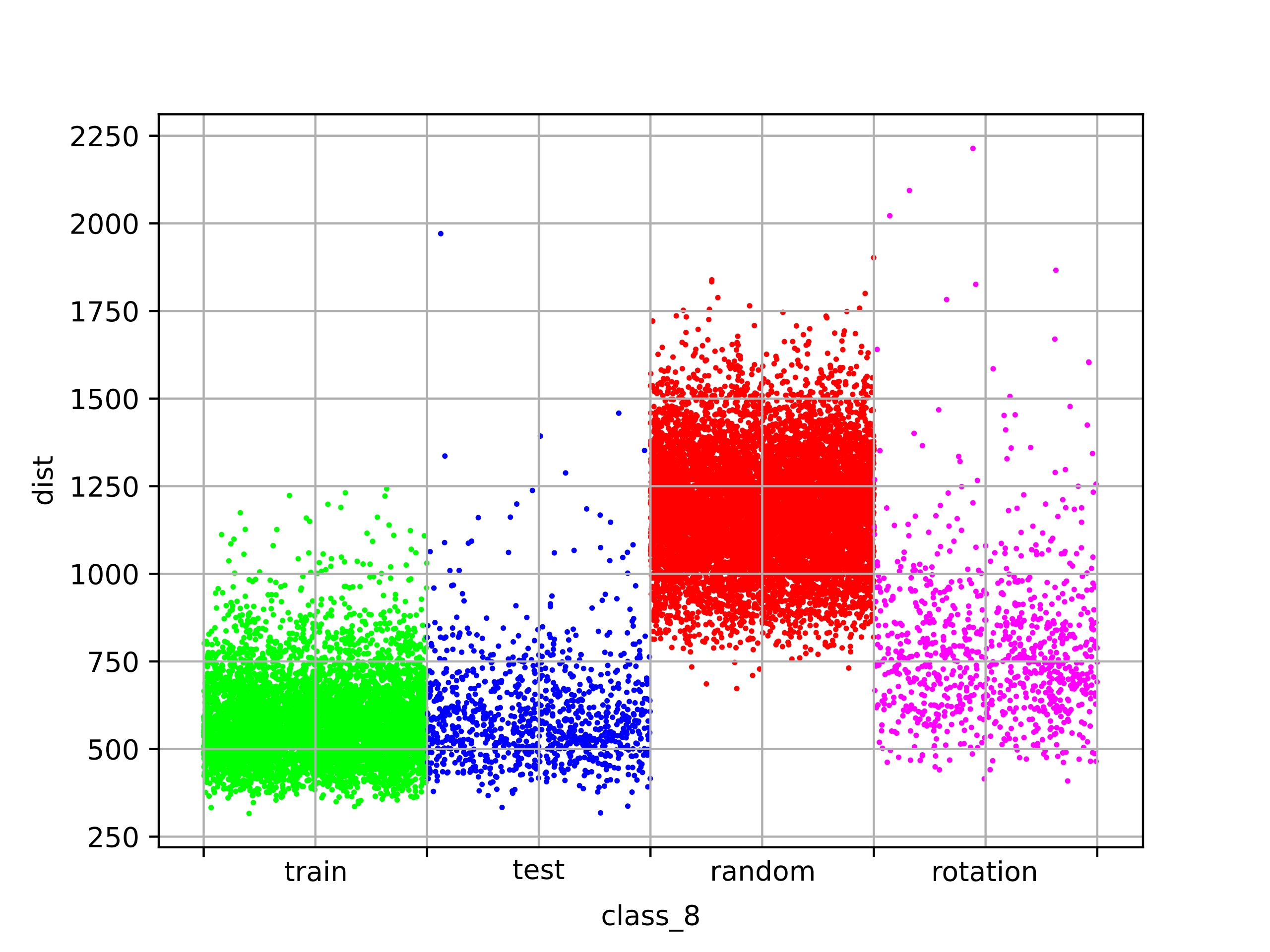} &
  \includegraphics[keepaspectratio=1,width=0.5\linewidth]{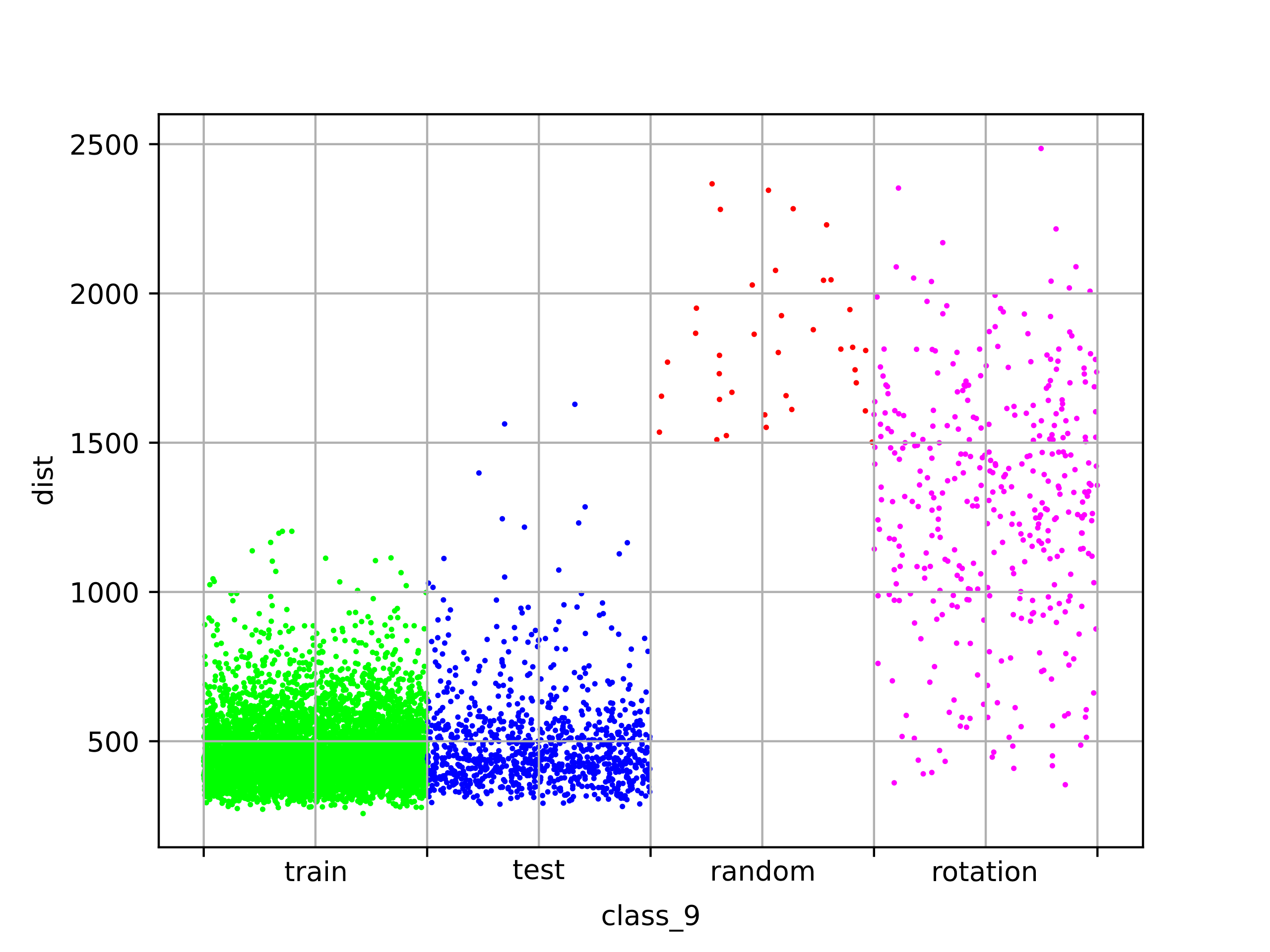}\\
\end{tabular}
\end{center}
\caption{Distance Distributions}
\label{fig-dist-distr}
\end{figure}

Figure \ref{fig-dist-distr} shows the distance plots of the compared categories, namely training (green), test (blue), random (red), and rotation (magenta). Distances have been computed using Equation \ref{eq-prob:e}.

As mentioned in Section \ref{lab-non-param}, a small distance means that the input is likely to have a computational profile close to those observed during training for the same class, while a high distance for an input says that its computational profile is far from those observed during training.

It can be seen in some of the figures, specifically classes 1, 7, and 9, that both the random and rotation likelihood are somehow separated from the train and test sets, the noise being a bit farther away. However, for the rest of the figures, and most of the classes, the rotations likelihood distribution is in fact well spread along a wide range and is in line with the train and test sets in some cases. As for the random set, it remains mostly separated from the train and test. This applies to the classes 0, 2, 3, 4, 5, 6, and 8. This shows that in most cases, the rotations images end up overlapping with the distances of the train and test set, while the random images remain separated.

Distribution parameters for all MNIST classes are reported in Tables \ref{tab-train-dist-params}, \ref{tab-test-dist-params}, \ref{tab-rnd-dist-params}, and \ref{tab-rot3-dist-params}, for the different categories.
It can be remarked that the class distributions are quite different. Classes 3, 5 and 7 are very robust with respect to noise, while many noisy inputs fall into classes 6 and 8.

\begin{table}[htbp]
\setlength\tabcolsep{5pt}
\renewcommand{\arraystretch}{1.3}
\begin{center}
\begin{tabular}{||c|c|c|c||}
\hline
class & \# of Images & Avg Distance & Std Deviation\\
\hline
\hline
0 & 5966 & 457.5 & 207.9 \\
\hline
1 & 5999 & 534.1 & 141.8 \\
\hline
2 & 5906 & 437.3 & 173.1 \\
\hline
3 & 5989 & 532.5 & 138.9 \\
\hline
4 & 5913 & 453.3 & 159.4 \\
\hline
5 & 5997 & 594.8 & 131.6 \\
\hline
6 & 5836 & 493.2 & 179.4 \\
\hline
7 & 5995 & 487.4 & 143.5 \\
\hline
8 & 6000 & 578.7 & 119.5 \\
\hline
9 & 5861 & 469 & 115.3 \\
\hline
\hline
\end{tabular}
\end{center}
\vspace*{1em}
\caption{Distance Distribution Parameters for Train}
\label{tab-train-dist-params}
\end{table}

\begin{table}[htbp]
\setlength\tabcolsep{5pt}
\renewcommand{\arraystretch}{1.3}
\begin{center}
\begin{tabular}{||c|c|c|c||}
\hline
class & \# of Images & Avg Distance & Std Deviation\\
\hline
\hline
0 & 889 &  480.8 & 288.5 \\
\hline
1 & 984 & 567.2 & 229.7 \\
\hline
2 & 873 & 457.1 & 212.2 \\
\hline
3 & 924 & 571.3 & 195.2 \\
\hline
4 & 877 & 476.5 & 190.8 \\
\hline
5 & 987 & 633.1 & 155.3 \\
\hline
6 & 735 & 506.4 & 214.8 \\
\hline
7 & 986 & 510.5 & 191.2 \\
\hline
8 & 990 & 605.6 & 155 \\
\hline
9 & 947 &  497 & 161.2 \\
\hline
\hline
\end{tabular}
\end{center}
\vspace*{1em}
\caption{Distance Distribution Parameters for Test}
\label{tab-test-dist-params}
\end{table}

\begin{table}[htbp]
\setlength\tabcolsep{5pt}
\renewcommand{\arraystretch}{1.3}
\begin{center}
\begin{tabular}{||c|c|c|c||}
\hline
class & \# of Images & Avg Distance & Std Deviation\\
\hline
\hline
0 & 1420 & 1702 & 250.3 \\
\hline
1 & 601 & 2657.9 & 294.1 \\
\hline
2 & 161 & 1603.5 & 239 \\
\hline
3 & 9 & 1650.6 & 131 \\
\hline
4 & 925 & 1756.1 & 265.2 \\
\hline
5 & 0 & - & - \\
\hline
6 & 47203 & 1333.5 & 198.9 \\
\hline
7 & 0 & - & - \\
\hline
8 & 9645 & 1180.1 & 159.7 \\
\hline
9 & 36 & 1838.2 & 243.7 \\
\hline
\hline
\end{tabular}
\end{center}
\vspace*{1em}
\caption{Distance Distribution Parameters for Random}
\label{tab-rnd-dist-params}
\end{table}

\begin{table}[htbp]
\setlength\tabcolsep{5pt}
\renewcommand{\arraystretch}{1.3}
\begin{center}
\begin{tabular}{||c|c|c|c||}
\hline
class & \# of Images & Avg Distance & Std Deviation\\
\hline
\hline
0 & 2751 & 1366.4 & 571.7 \\
\hline
1 & 599 & 1673.2 & 399 \\
\hline
2 & 746 & 642.3 & 385.7 \\
\hline
3 & 1310 & 934.1 & 385.7 \\
\hline
4 & 335 & 870 & 521.9 \\
\hline
5 & 483 & 675.4 & 179.6 \\
\hline
6 & 1247 & 925.2 & 413.3 \\
\hline
7 & 1166 & 1292.2 & 420 \\
\hline
8 & 910 & 793.7 & 218.1 \\
\hline
9 & 373 & 1285.3 & 403.4 \\
\hline
\hline
\end{tabular}
\end{center}
\vspace*{1em}
\caption{Distance Distribution Parameters for Rotations}
\label{tab-rot3-dist-params}
\end{table}

We have also performed two by two 2-sample statistical tests, namely Epps-Singleton and Cliff's delta \cite{scipi_stats}. These were done to compare the Training/Random image sets (Table \ref{tab:stat-train-rnd}), the Training/Rotation image sets (Table \ref{tab:stat-train-rot3}), and the Training/Test image sets for comparison (Table \ref{tab:stat-train-test}).

As expected, the Training/Test sets produce results that show that the images are not necessarily from different distributions. The sig- and p- values are relatively large, and the Cliff's delta is negligible for all classes. In contrast, Training/Random and Training/Rotation distribution comparisons show that these distributions are quite different, since Cliff's delta is always large, except for rotated classes 2 and 4 that sport medium delta and class 5 for which delta is small.
Additionally, we have also performed the following two by two 2-sample statistical tests: Anderson, Kolmogorov-Smirnov, Wilcoxon\cite{scipi_stats}, and Cohen-D that was implemented by ourselves. Results were consistent and similar to those obtained for Epps-Singleton and Cliff's delta, but have not been reported, because they were somewhat redundant.
As for the Training/Random and Training/Rotation comparisons, the sig- and p- values are much smaller and inferior to the Bonferroni's correction threshold 'alpha' = 0.005 described in Section \ref{lab-stats}.

Also as expected, since there is much more distance between the likelihood of Training/Random, the Cliff's delta are all classified as large, whereas for Training/Rotation, most are large with few medium and one small delta, showing that rotations likelihood lie in an extended range with respect to training, tests, and noise.

The presented approach allows the independent assessment of different classes. Indeed, as it can be seen by looking at Figure \ref{fig-dist-distr} and Table \ref{tab-rot3-dist-params}, error rotation transformations that are best classified into class 5 are very similar in distribution to training and test. For this class, likelihoods of error rotations are statistically indistinguishable from cases in normal training or tests. Cliff's delta is equal to -0.256 and ``small''. In contrast, error rotations classified into class 1 are fairly separable from training cases and Cliff's delta is equal to -0.994 and ``large''.

\begin{table}[htbp]
\setlength\tabcolsep{3pt}
\renewcommand{\arraystretch}{1.3}
\begin{center}
\begin{tabular}{||c|c|c|c|c||}
\hline
\hline
& \multicolumn{2}{c|}{Epps-Singleton} & \multicolumn{2}{c|}{Cliff's} \\
\hline
class & statistic & p-value & delta & size \\
\hline
\hline
0 & 30.9 & 3.2e-06 & 0.0158 & negligible \\
\hline
1 & 14.1 & 0.007 & -0.0388 & negligible \\
\hline
2 & 11.4 & 0.02 & -0.0269 & negligible \\
\hline
3 & 40.9 & 2.8 & -0.0809 & negligible \\
\hline
4 & 16.1 & 0.003 & -0.0474 & negligible \\
\hline
5 & 59.5 & 3.8e-12 & -0.139 & negligible \\
\hline
6 & 25.3 & 4.4e-05 & 0.00314 & negligible \\
\hline
7 & 9.2 & 0.06 & -0.0428 & negligible \\
\hline
8 & 25 & 5.1e-05 & -0.0820 & negligible \\
\hline
9 & 28.1 & 1.2e-05 & -0.0698 & negligible \\
\hline
\hline
\end{tabular}
\end{center}
\vspace*{1em}
\caption{Statistical 2-Sample Tests: training / test}
\label{tab:stat-train-test}
\end{table}

\begin{table}[htbp]
\setlength\tabcolsep{3pt}
\renewcommand{\arraystretch}{1.3}
\begin{center}
\begin{tabular}{||c|c|c|c|c||}
\hline
\hline
& \multicolumn{2}{c|}{Epps-Singleton} & \multicolumn{2}{c|}{Cliff's} \\
\hline
class & statistic & p-value & delta & size \\
\hline
\hline
0 & 166259.54 & 0.0 & -0.998 & large \\
\hline
1 & 3828.08 & 0.0 & -1 & large \\
\hline
2 & 790.16 & 1.04e-169 & -0.999 & large \\
\hline
3 & 229.90 & 1.39e-48 & -1.0 & large \\
\hline
4 & 22680.87 & 0.0 & -1.0 & large \\
\hline
5 & - & - & - & - \\
\hline
6 & 318458.42 & 0.0 & -0.994 & large \\
\hline
7 & - & - & - & - \\
\hline
8 & 141370.12 & 0.0 & -0.993 & large \\
\hline
9 & 94.05 & 1.82e-19 & -1.0 & large \\
\hline
\hline
\end{tabular}
\end{center}
\vspace*{1em}
\caption{Statistical 2-Sample Tests: training / random}
\label{tab:stat-train-rnd}
\end{table}

\begin{table}[htbp]
\setlength\tabcolsep{3pt}
\renewcommand{\arraystretch}{1.3}
\begin{center}
\begin{tabular}{||c|c|c|c|c||}
\hline
\hline
& \multicolumn{2}{c|}{Epps-Singleton} & \multicolumn{2}{c|}{Cliff's} \\
\hline
class & statistic & p-value & delta & size \\
\hline
\hline
0 & 11339.3 & 0.0 & -0.875 & large \\
\hline
1 & 969 & 1.9e-208 & -0.994 & large \\
\hline
2 & 274.2 & 4e-58 & -0.409 & medium \\
\hline
3 & 2987.0 & 0.0 & -0.747 & large \\
\hline
4 & 184.1 & 9.8e-39 & -0.455 & medium \\
\hline
5 & 112.1 & 2.6e-23 & -0.256 & small \\
\hline
6 & 1998.4 & 0.0 & -0.674 & large \\
\hline
7 & 7651.0 & 0.0 & -0.954 & large \\
\hline
8 & 1624.3 & 0.0 & -0.674 & large \\
\hline
9 & 620.3 & 6.2e-133 & -0.934 & large \\
\hline
\hline
\end{tabular}
\end{center}
\vspace*{1em}
\caption{Statistical 2-Sample Tests: training / rotation}
\label{tab:stat-train-rot3}
\end{table}

\section{Future Research}
\label{lab-future-res}

In Section \ref{lab-affine-tranf}, we describe multiple types of affine transformations, as also reported in Table \ref{tab:affine_trans}. However, we focused on rotation for the purposes of the presented experiments. Further research could be performed in the future on the other types of affine transformations and their parameters, for the purposes of comparison.

 Also, considering that the image generation process involves randomness, the experiments should be repeated several times with different generated images to avoid coincidental results. Furthermore, repeating these experiments should allow more insight into the reasons behind why certain classes seem to be assigned most noisy (random) images, while some other classes have very little or no assignment at all. This outcome is quite surprising, and it would be interesting to investigate it further.

It would also be interesting to investigate whether the estimated density probabilties of neuron activation levels suffer from outliers that would contribute to the non-normality, skewness and sparsity of the observed distributions. Robust statistics approaches \cite{robust_stats_histogram_ieee_2000} may be used to investigate anomalies and outliers.

Further investigation is also needed to understand what are the attributes or features that make certain classes more sensitive to the presented rotation-based adversarial attacks.

By speculation we may think of domain characteristics that would make, for example, a rotated tee-shirt look like a shirt, and so on. Additional experiments should be run on other affine tranformations or combinations of them to study whether this happens for other transformations, too. Also, more experiments should be performed by using additional networks inferred with different initial weights and different hyperparameters, to assess the impact of class sensitivity to adversarial inputs.

\section{Related Work}
\label{lab-rel-work}

Some sort of computational profile likelihood has been used in approaches to detect or defend against adversarial attacks. For example, Papernot \cite{papernot_distillation_arxiv_2015, papernot_sp_2016} presented distillation of neuron activation levels from training and adversarial inputs and training a secondary DNN as a defensive measure for robustness. Another paper \cite{papernot_knn_arxiv_2018} investigated neuron activation levels and a secondary classifier based on nearest neighbors computed using locality-sensitive hashing.

We share the extraction of activation levels, but we model them with non-parametric likelihood of an input given a class, without the need of secondary training.

Kim \cite{miryung_fse_2020} recently showed that neuron coverage is not significantly statistically related to adversarial case detection. Indeed, we consider actual neuron activation levels as some sort of statistical signature used to measure the ``reasoning'' likelihood of computational profiles during prediction.

SADL approach \cite{surprise_icse_2019} uses neuron activation values to calculate the level of ``surprise" between training images and adversarial images. They then use these ``surprise" values to retrain a classifier to avoid the misclassification of those adversarial images.

RAID \cite{raid_arxiv_2020} considers the activation levels of a subset of relevant neurons with the highest difference values with respect to adversarial inputs to retrain a secondary classifier and assess the confidence of the predictions.

We considered only the last but one layer in the presented experiments as discussed in Section \ref{lab-ttv}. In principle, this would allow a faster computation, since we consider less neurons and have higher statistical independence from previous layers, in contrast with analyzing the whole network \cite{papernot_knn_arxiv_2018, surprise_icse_2019, raid_arxiv_2020}.

Other approaches use activation levels to detect adversarial techniques. Some examples are \cite{gong_twins_arxiv_2017, grosse_stat_detection_arxiv_2017, metzen_iclr_2017}. They all demonstrate satisfactory results and, very similarly to RAID \cite{raid_arxiv_2020}, they all rely on training a secondary classifier to be effective, while our approach does not need secondary training.

Other approaches to detect adversarial images vary from using techniques like Principal Component Analysis \cite{Bhagoji_dimensionality_RA_2016, hendrycks_early_adv_arxiv_2017, Li_conv_filter_stats_arxiv_2016} to Kernel-Density Estimation and Bayesian Neural-Network Uncertainty \cite{feinman_detecting_adv_arxiv_2017}. However each and every one of these techniques can be defeated by choosing a specific loss function depending on the defense \cite{carlini_bypass_adv_detection_arxiv_2017}.

Ravi Mangal et al. \cite{orso_nier_2019} presented an approach for robustness of neural networks based on non-adversarial real-world input probability distributions. We share the perspective of stochastic modeling, but we concentrate on computational profiles of networks rather than input data distributions. 

A good review of adversarial attacks can be found in the NISTIR 8269 report \cite{taxonomy_nist_2019}, where several different adversarial attacks and categories of attacks are described. According to such a report, Data Access Attacks are attacks in which the adversary uses all or part of the training data to create and train a new model that they then use to evaluate their perturbed images. When it comes to evasion attacks, there are the Fast Gradient Sign Method, Jacobian-Based Saliency Map Attack, and Limited memory Broyden–Fletcher–Goldfarb–Shanno (BFGS) method \cite{attacks_akhtar_ieee_2018, papernot_sok_ieee_2018, liu_bfgs_mathprog_1989}. In these cases, some advanced computations are performed in an optimization exercise to look at what perturbations to the image would cause a significant change in the model's cost function. Other types of adversarial attacks also include (but are not limited to) Poisoning Attacks (in which the adversary changes the input data or model directly), Data Injection (in which the adversary injects new data into the original set of training inputs, and Data Manipulation (in which the adversary manipulates the existing training images or labels).
Further research could include comparing the likelihood distributions of our metamorphic test cases to those of other adversarial attacks reported in the literature.

The analysis of a network's internal computation values to differentiate between adversarial samples and benign inputs has been explored and presented in the literature \cite{dnn_invar_shiqing_ndss_2019, dnn_mutation_wang_icse_2019}. This presented paper extends the investigation to include metamorphic tests.

\section{Threats To Validity}
\label{lab-ttv}

Our experiments have been performed on one network trained on MNIST images. A recent paper \cite{lin_tan_ase_2020} showed a high variability across different trained networks. In this perspective, presented results may not be the same across different architectures trained on the same data. Although we didn't perform variability analysis, this can be done in further studies to measure and assess the impact of variability on the likelihood of metamorphic test cases across architectures. Furthermore, to mitigate the risk of the bias involved for randomized tasks that use random seeds, we input a random seed and in turn used that random seed to generate another random seed that was then used throughout the experiment. This ensures that the seed being used in the training process is not directly chosen by us.
Additionally, presented results on random-pixel images depend on the choice of random generator, the design of filling an image by rows or by columns, and the choice of not re-initializing random seeds between images. Although different choices may produce slighly different results, we do not believe this is an issue, but multiple and different random generation schemes may be deployed to reduce the risks of dependency on the random generation process.

When classifying samples from the random image set, we noticed that the network results were fairly unbalanced. We observed many more samples in some classes such as class 6 and 8 than in others. The inferred network from training may be biased towards those classes, and this may have influenced our results. Multiple networks inferred using different initial values or hyperparameters could be used to investigate the extent of this bias. It could be also related to the method in which we generated the random images. The reasons for these unbalanced results have not been investigated in this paper and are left to further research.

In this paper, we didn't compare histograms vs. Gaussian mixtures density estimators \cite{neural_prob_zheng_nips_2018}. We believe that the two approaches may be used as alternatives. Intuitively, we chose histograms because of the very highly skewed and highly sparse distribution of probability densities observed in our data. Other sets of data or domains may have different properties.

Also, our experiments have been performed on one large database of clothes images \cite{MNIST-fashion}. Presented results may not be generalizable to other datasets and other domains. Additional experiments have to be repeated and performed.

We considered neurons in the last but one layer only. The literature \cite{surprise_icse_2019} suggests that layer sensitivity varies: ``surprise adequacy (SA)'' is more effective on the final (surface) layers of a network than the initial (deepest) layers for the MNIST dataset. Conversely, SA seems more effective in deepest layers for the CIFAR dataset. Possibly, our presented results may not generalize on other datasets and more layers should be investigated to compute the proposed likelihood.

In addition, we experimented on architectures that present fully connected output layers with the RELU transfer function. Although these architectures are quite common and frequent, obtained results are not generalizable to different structures and additional experiments have to be repeated and performed.

\section{Conclusion}
\label{lab-concl}

We have investigated the distribution of ``reasoning'' likelihood based on computational profiles of metamorphic test cases with respect to the likelihood distributions of training, test and error control cases.

Likelihood has been computed using non-parametric estimation of neuron activation levels probabilities under the hypothesis of each distinct output class. Estimation was performed using normal training data only, without additional knowledge or secondary training about metamorphic data, features, or behavior.

Experiments on images from the MNIST-fashion database for training and tests, on images composed of random pixels for error control-data, and on metamorphically generated corner case images have been performed by computing the likelihood of the best predicted class.

Training and test sets' activation patterns are very similar to one another. On the other hand, the activation patterns of totally randomized inputs are substantially different. For the metamorphic test sets the activation patterns cover an interval that includes that of training/test and more.

Somehow, metamorphic testing can be seen as exercising the network paths and profiles in regions excluded from random testing.

Metamorphic test cases cover a wide range of likelihood and thus, when filtered to consider only the training/test likelihood range, can be used as additional aggressive test cases or could even be considered as adversarial attacks that evade defenses based on ``computational profile'' likelihood computation.

Although experimented on image recognition, the presented approach is definitely not limited to image recognition, but could be applied to all networks that have fully-connected and soft-max final layers.

\ifCLASSOPTIONcompsoc
  \section*{Acknowledgments}
\else
  \section*{Acknowledgments}
\fi

The authors wish to thank the revievers for their constructive comments and suggestions.
They also wish to thank the DEEL projet \mbox{CRDPJ-537462-18} funded by the National Science and Engineering Research Council of Canada (NSERC) and the Consortium for Research and Innovation in Aerospace in Québec (CRIAQ), together with its industrial partners Thales Canada Inc., Bell Textron Canada Ltd., CAE Inc. and Bombardier Inc.

\balance
\bibliographystyle{IEEEtran}
\bibliography{IEEEabrv,currentDnn}

\end{document}